\documentclass[acmtog, authorversion]{acmart}

\AtBeginDocument{%
  \providecommand\BibTeX{{%
    \normalfont B\kern-0.5em{\scshape i\kern-0.25em b}\kern-0.8em\TeX}}}

\setcopyright{acmcopyright}
\copyrightyear{2018}
\acmYear{2018}
\acmDOI{XXXXXXX.XXXXXXX}

\acmConference[Conference acronym 'XX]{Make sure to enter the correct
  conference title from your rights confirmation emai}{June 03--05,
  2018}{Woodstock, NY}
\acmBooktitle{Woodstock '18: ACM Symposium on Neural Gaze Detection,
 June 03--05, 2018, Woodstock, NY} 
\acmPrice{15.00}
\acmISBN{978-1-4503-XXXX-X/18/06}

\usepackage{multirow}
\usepackage{subcaption}
\usepackage{float}
\usepackage{graphicx}
\usepackage{dsfont}
\begin{document}

\title{Uncertainty Quantification in Inverse Models in Hydrology}


\author{Somya Sharma}
\affiliation{%
 \institution{University of Minnesota - Twin Cities}
 \city{Minneapolis}
 \state{MN}
 \country{USA}}

 \author{Rahul Ghosh}
\affiliation{%
 \institution{University of Minnesota - Twin Cities}
 \city{Minneapolis}
 \state{MN}
 \country{USA}}
 
 \author{Arvind Renganathan}
\affiliation{%
 \institution{University of Minnesota - Twin Cities}
 \city{Minneapolis}
 \state{MN}
 \country{USA}}

 \author{Xiang Li}
\affiliation{%
 \institution{University of Minnesota - Twin Cities}
 \city{Minneapolis}
 \state{MN}
 \country{USA}}

\author{Snigdhansu Chatterjee}
\affiliation{%
 \institution{University of Minnesota - Twin Cities}
 \city{Minneapolis}
 \state{MN}
 \country{USA}}
 
 \author{John Nieber}
\affiliation{%
 \institution{University of Minnesota - Twin Cities}
 \city{Minneapolis}
 \state{MN}
 \country{USA}}
 
 \author{Christopher Duffy}
\affiliation{%
 \institution{Pennsylvania State University}
 \city{University Park}
 \state{PA}
 \country{USA}}
 \author{Vipin Kumar}
\affiliation{%
 \institution{University of Minnesota - Twin Cities}
 \city{Minneapolis}
 \state{MN}
 \country{USA}}

\renewcommand{\shortauthors}{Sharma and Kumar}

\begin{abstract}
  In hydrology, modeling streamflow remains a challenging task due to the limited availability of basin characteristics information such as soil geology and geomorphology. These characteristics may be noisy due to measurement errors or may be missing altogether. To overcome this challenge, we propose a knowledge-guided, probabilistic inverse modeling method for recovering physical characteristics from streamflow and weather data, which are more readily available. We compare our framework with state-of-the-art inverse models for estimating river basin characteristics. We also show that these estimates offer improvement in streamflow modeling as opposed to using the original basin characteristic values. Our inverse model offers 3\% improvement in R$^2$ for the inverse model (basin characteristic estimation) and 6\% for the forward model (streamflow prediction). Our framework also offers improved explainability since it can quantify uncertainty in both the inverse and the forward model. Uncertainty quantification plays a pivotal role in improving the explainability of machine learning models by providing additional insights into the reliability and limitations of model predictions. In our analysis, we assess the quality of the uncertainty estimates. Compared to baseline uncertainty quantification methods, our framework offers 10\% improvement in the dispersion of epistemic uncertainty and 13\% improvement in coverage rate. This information can help stakeholders understand the level of uncertainty associated with the predictions and provide a more comprehensive view of the potential outcomes.
\end{abstract}





\maketitle

\section{Introduction}

Researchers in scientific communities study engineered or natural systems and their responses to external drivers. In hydrology, streamflow prediction \cite{ghimire2021streamflow, ghosh2022robust} is one crucial research problem for understanding hydrology cycles, flood mapping, water supply management, and other operational decisions. For a given entity (river basin/catchment), the response (streamflow) is governed by external drivers (meteorological data) and complex physical processes specific to each entity (basin/entity characteristics). Machine learning (ML) paradigms in inverse modeling enable us to infer entity characteristics from streamflow response. In our study, for the same amount of precipitation (external driver), two river basins (entities) can have very different streamflow (response) values depending on their soil geology (entity characteristic)~\cite{newman2015gridded} - this presents the issue of navigating a large search space to learn one of the many right model structures. Recently, Knowledge-guided self-supervised learning (KGSSL) ~\cite{ghosh2022robust}, the state-of-the-art inverse model for extracting these entity characteristics, was proposed. The framework uses a self-supervised learning paradigm, where ML models are trained using labels that can be generated without any external annotation process. However, it is not capable of quantifying uncertainty. This may present challenges in its adoption for real-life decision-making. This is because, in hydrology, observed data are not only impacted by measurement uncertainty in static characteristics, arising from measurement errors and use of estimation methods, but may further be affected by uncertainties arising from hydrological approximations, weather-forecasting based distributional shifts, and dam regulations. 

Developing inverse models that can quantify uncertainty requires addressing several challenges. Often, the measured characteristics are only surrogate variables for the actual entity characteristics, leading to inconsistencies and high uncertainty \cite{Addor2017}. Moreover, in real-world applications, these characteristics may be essential in modeling the driver-response relation. However, they may be completely unknown, not well understood, or not present in the available set of entity characteristics \cite{ghosh2022robust}. A principled method of managing this uncertainty due to imperfect data can contribute to improving trust in data-driven decision-making from these methods.

In this paper, we introduce uncertainty quantification in learning representations of static characteristics. Such a framework complements explainability efforts by providing additional context and insights into the reliability, limitations, and decision-making process of machine learning models. For instance, the Equifinality of hydrological modeling (different model representation result in the same model results) is a widely known phenomenon affecting the adoption of hydrology models in practice \cite{her2019uncertainty}. Uncertainty in model structure and input data is also widespread. Studying the effect of such challenges can help improve the trust of water managers, improve process understanding, reduce costs, and make predictions explainable \cite{mcmillan2018hydrological}. 

To achieve this, we propose a Bayesian inverse model for simultaneously learning representations of static characteristics and quantifying uncertainty in these predictions. As a consequence, we analyze the framework's reconstruction capabilities. We modify the KGSSL autoencoder architecture such that the parameters in the encoder are estimated using the \textit{Bayes by Backprop} weight perturbation method. This enables learning of the posterior weight distribution and uncertainty quantification in static characteristic reconstructions. We also propose an uncertainty-based learning (UBL) method to reduce epistemic uncertainty (uncertainty in predictions due to imperfect models and imperfect data) in our reconstructions. This method utilizes a spectral regularization-based objective formulation wherein reconstructions with higher uncertainty are penalized in the loss. We also demonstrate the improvement in streamflow prediction (in the forward model) using these robust reconstructed static characteristics (6\% increase in test $R^2$). We provide model performance for inverse and forward models and compare it against the baselines, KGSSL \cite{ghosh2022robust} and CT-LSTM \cite{kratzert2019towards}, both state-of-the-art frameworks for streamflow modeling. We also compute the coverage rate of how often the observed values lie within the bounds of the inferred static characteristics' posterior prediction distribution. In practice, this can help water managers and the public to understand if we can reliably obtain a close enough prediction, even if we are not always accurate - analysis that can not be done with the deterministic inverse model. UBL offers a 4\% increase in coverage rate.

\vspace{-10pt}

\section{Related Work}


In hydrology, river flow modeling is a well-studied problem, with several recent advances focusing on using ML methods to build streamflow prediction models \cite{kratzert2019towards}. However, estimating the inverse mapping from streamflow to river basin characteristics remains less explored. Due to the problem of equifinality, the estimation of river basin characteristics still remains a challenging task. In physical sciences \cite{ woolway2021winter, pecha2021determination, dao2021improving}, several recent advances have focused on solving inverse problems. Unlike standard inversion methods in mathematics, which rely on non-linear optimization for calculating the inverse of a forward model, recent machine learning methods allow us to learn the inverse mapping from datasets. This makes it imperative to mitigate any representation error and data biases before solving the inverse problem \cite{asim2020invertible}. Further, within this vast array of methodologies, the selection of the right method is crucial - since searching for an inverse mapping may be difficult due to the large search space. Bayesian optimization and iterative gradient descent-based methods may only provide a locally optimal inverse map \cite{lavin2021simulation}. Therefore, a principled MAP formulation and generative modeling may be viable for addressing these data-related issues \cite{sun2020deep, whang2021solving}. Moreover, inverse modeling approaches that rely on a single neural network may not accurately capture the spatio-temporal heterogeneity in basin characteristics. Also, entity characteristics can be unknown or noisy (due to measurement or estimation bias). Therefore, a robust framework for learning entity characteristics can be useful in hydrology. A recent study proved the efficacy of self-supervised autoencoder-based machinery for recovering static characteristics from streamflow values \cite{ghosh2022robust}. However, due to the measurement uncertainty and hydrological uncertainty in the input data, it is difficult to evaluate if the predictions from such models are trustworthy. In such a case, using generative models for uncertainty quantification not only allows for a complete recovery of the entity characteristics distribution but also allows us to evaluate uncertainty arising from different sources within the framework \cite{asim2020invertible, daw2021pid, whang2021solving}. 

We develop a Bayesian inverse model for robust recovery of the complete distribution of the entity characteristics. Our framework achieves this by obtaining estimates of static variables from time series driver-response data. This, however, introduces temporal bias in our static characteristics. We also propose an uncertainty-based learning scheme to reduce the uncertainty associated with this temporal bias in inverse model estimates of static characteristics. 

\vspace{-5pt}

\section{Method}
\subsection{Inverse Model}

\begin{figure}
    \centering
    \includegraphics[scale=0.27]{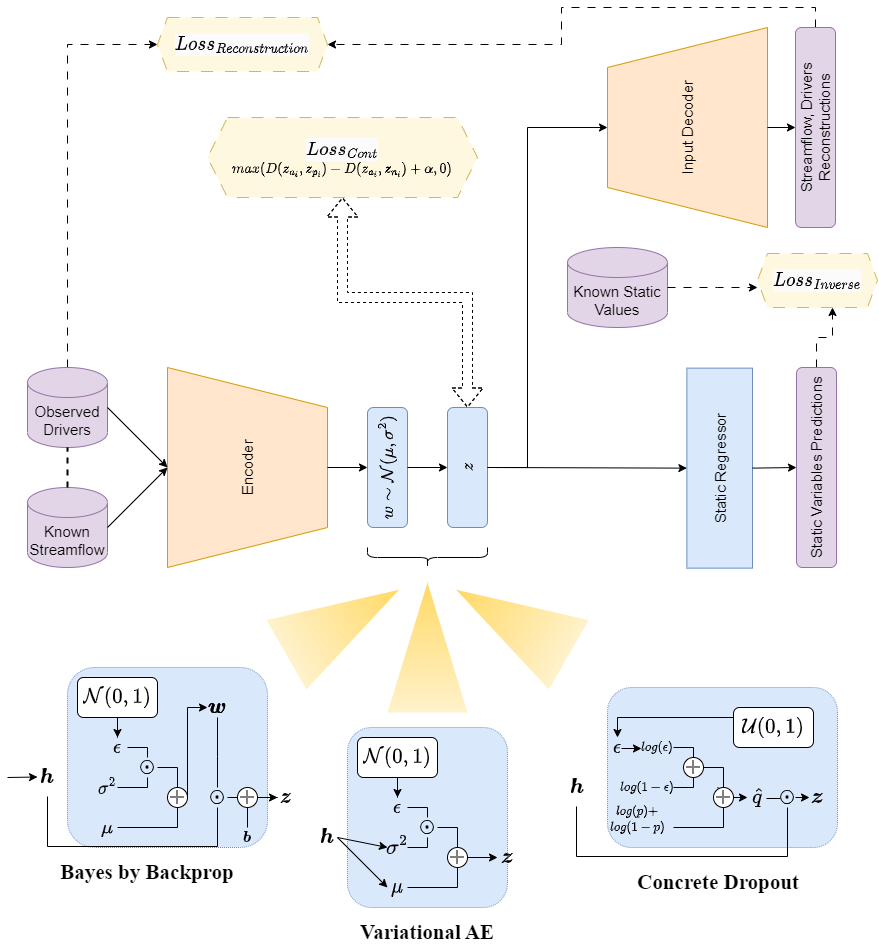}
    \caption{Bayesian Inverse Model (BIM)}
    \label{fig:bim}
\vspace{-20pt}
\end{figure}

Streamflow dynamics can vary widely depending on the inherent basin-level static characteristics. Motivated by the recent success of autoencoders in estimating the static characteristics information from streamflow \cite{ghosh2022robust}, we also incorporate an autoencoder based inverse model for learning basin characteristics. In our problem setting, each river basin , $i$'s ($i=1,...,N$), weather drivers ,$x_i^j \in \mathbb{R}^{\mathcal{D}_x}$, and streamflow data, $y_i^j \in \mathbb{R}^{\mathcal{D}_y}$ can be leveraged to learn an inverse mapping to the static characteristics, $z_i^j \in \mathbb{R}^{\mathcal{D}_z}$ at the $j^{th}$ time step \cite{ghosh2022robust}. The Sequence Encoder, comprised of a bidirectional LSTM, encodes the driver-response time-series. Each (forward and backward) LSTM use $[\boldsymbol{x^t};y^t]$ input to generate the carry state and the hidden state $h = [h_{\text{forward}}; h_{\text{backward}}]$. Using a ReLU tranformation , a linear layer is used in the encoder to get a transformation of the hidden embedding. These transformed embeddings are used as input to the LSTM decoder $\mathcal{D}$. The observed sequence $\mathcal{S}_{e_i}$ are compared with the reconstructed sequence $\hat{\mathcal{S}}_{e_i}$ from the decoder in the reconstruction loss, $\mathcal{L}_{Rec} = \frac{1}{2N} \sum_{e\in \{a,p\}} \sum_{i=1}^N MSE(\hat{S}_{e_i}, S_{e_i})$.

\begin{equation}
\small
    \begin{split}
        \boldsymbol{i_t}    &= \sigma (\boldsymbol{W_i}\left[[\boldsymbol{x^t};y^t];\boldsymbol{h^{t-1}}\right] + \boldsymbol{b_i})\\
        \boldsymbol{f_t}    &= \sigma (\boldsymbol{W_f}\left[[\boldsymbol{x^t};y^t];\boldsymbol{h^{t-1}}\right] + \boldsymbol{b_f})\\
        \boldsymbol{g_t}    &= \sigma (\boldsymbol{W_g}\left[[\boldsymbol{x^t};y^t];\boldsymbol{h^{t-1}}\right] + \boldsymbol{b_g})\\
        \boldsymbol{o_t}    &= \sigma (\boldsymbol{W_o}\left[[\boldsymbol{x^t};y^t];\boldsymbol{h^{t-1}}\right] + \boldsymbol{b_o})\\
        \boldsymbol{c_t}    &= \boldsymbol{f_t} \odot \boldsymbol{c_{t-1}} + \boldsymbol{i} \odot \boldsymbol{g_t}\\
        \boldsymbol{h_t}    &= \boldsymbol{o_t} \odot \tanh{(\boldsymbol{c_t})}\\
    \end{split}
\end{equation}

The spatial heterogeneity among different river basins can further be leveraged to learn the differences in basin characteristics. Knowledge-guided Contrastive Loss ensures that the inherent hydrological and physical association among similar entities can allow for more efficient representation learning. The implicit physical properties (in embeddings $h_{a_i}$ and $h_{b_i}$) of ``positive pairs" of sequences ($S_{a_i}$ and $S_{p_i}$, respectively) are compared to other entity sequences. Here, positive pairs (of sequences) refers to learning from temporal associations in the basin while negative pairs (of basins) refers to samples that enable learning from spatial correlation among basins.

\begin{equation}
\small
    \begin{split}
        l(a_i,p_i) = & \frac{\exp{(sim(\boldsymbol{h_{a_i}}, \boldsymbol{h_{p_i}})/\tau)}}{\sum_{e\in\{a,p\}}\sum_{j=1}^N\exp{(sim(\boldsymbol{h_{a_i}}, \boldsymbol{h_{e_j}})/\tau)}}\\
        + &\frac{\exp{(sim(\boldsymbol{h_{p_i}}, \boldsymbol{h_{a_i}})/\tau)}}{\sum_{e\in\{a,p\}}\sum_{j=1}^N\exp{(sim(\boldsymbol{h_{p_i}}, \boldsymbol{h_{e_j}})/\tau)}}
    \end{split}
\end{equation}
where, $sim(\boldsymbol{h_{a_i}}, \boldsymbol{h_{p_i}})=\frac{\boldsymbol{h_{a_i}}^T\boldsymbol{h_{p_i}}}{\|\boldsymbol{h_{a_i}}\|\|\boldsymbol{h_{p_i}}\|}$. Thus, the total contrastive loss for 2N such positive pairs is given as, $\mathcal{L}_{Cont} = \frac{1}{2N} \sum_{i=1}^N l(a_i,p_i)$.

$\mathcal{L}_{Cont}$ and $\mathcal{L}_{Rec}$ do not require any supervised information. This enables us to evaluate these losses on a large number of samples. Pseudo-Inverse Loss allows for a source of supervision to be based on the available static feature data. A feed-forward layer $I$ on sequence encoder output is used to estimate $\mathbf{\hat{z}} = I(\mathbf{h})$.

\begin{equation}
    \mathcal{L}_{Inv} = \frac{1}{N} \sum_{i=1}^N \frac{1}{z} \sum_{j=1}^z (z_i^j-\hat{z}_i^j)^2
\label{eq:Linv}
\end{equation}

Temporal heterogeneity in driver-response time series is a source of uncertainty in static feature reconstructions. For T time steps and W window size, $unc_i$ provides us with this standard deviation in static feature reconstruction over time, 


\begin{equation}
\label{eq:unc}
\small
    \boldsymbol{unc}_i = \sqrt{\frac{W}{T}\sum_{j=1}^{T/W} (\boldsymbol{\hat{z}}_i^j-\boldsymbol{\hat{z}}_i)^2}
\end{equation}

The objective function is 

\begin{equation}
    \label{eq:representation learning}
    \small
    \mathcal{L} = \lambda_1 \mathcal{L}_{Rec} + \lambda_2 \mathcal{L}_{Cont} + \lambda_3 \mathcal{L}_{Inv}
\end{equation}

where, reconstruction loss $\mathcal{L}_{Rec}$  enables accurate representation learning of $[x, y]$; contrastive loss $\mathcal{L}_{Cont}$ utilizes the implicit relationships among driver-response time series data, enabling invariant approximation of static features; pseudo-inverse loss (or static loss) $\mathcal{L}_{Inv}$ utilizes available static variable information to enable accurate representation learning. The loss coefficients are learned using hyper-parameter tuning.

\subsection{Uncertainty Estimation}
\label{sec:method-uncest}

The uncertainty in the estimation of static characteristics is obtained using a perturbation-based weight uncertainty method called \textit{Bayes by Backprop} \cite{blundell2015weight, wen2018flipout}. As a method that relies on learning the posterior distribution of weight parameters, \textit{Bayes by Backprop} can make different layers of the architecture non-deterministic. This allows us to measure and mitigate the uncertainty from different components incorporated in the framework. More recent studies also look at Bayesian deep learning models for their robustness properties \cite{carbone2020robustness, cardelli2019statistical,sharma2021winsorization}.

Introducing perturbations in weights while training has historically been used as a regularization method \cite{hanson1988comparing, srivastava2014dropout, kang2016shakeout, li2016whiteout,goodfellow2013maxout}. Some recent advances utilize perturbations to induce non-deterministic behavior in supervised learning models \cite{graves2011practical, wan2013regularization}. Several variations of Bayesian neural networks implement the reparameterization trick \cite{kingma2015variational} to learn affine transformation of perturbation using variational inference. All these methods rely on drawing a Gaussian perturbation term $\epsilon \sim \mathcal{N}(0, 1)$. The scale and shift parameters $\Sigma$ and $\mu$ can be learned by optimizing for variational free energy \cite{graves2011practical}. Therefore, the weight parameters, $w$, are learned as, $w = \mu + log(1+exp(\Sigma)) \odot \epsilon$. Here, $log(1+exp(\Sigma))$ is non-negative and differentiable. The variational parameters $\theta = \{\mu, \Sigma \}$ are minimized by variational free energy \cite{graves2011practical, friston2007variational, jaakkola2000bayesian, yedidia2000generalized, neal1998view} that ensures a trade-off between learning a complex representation of the data (the likelihood cost) and learning a parsimonious representation similar to the prior (complexity cost). The variational free energy cost \cite{blundell2015weight} can be written as,

\vspace{-5pt}

\begin{equation}
    \mathcal{F} = KL[q(w| \theta)|| Pr(w)] - \mathbb{E}_{q(w| \theta)}[log \hspace{3pt} Pr(\mathcal{D}|w)]
\end{equation}

The complexity cost is the KL divergence between the learned posterior distribution of weight parameters $q(w| \theta)$ and the prior probability $Pr(w)$. The likelihood cost includes the negative log likelihood indicating the probability that the weight parameters capture the complexity of the dataset $\mathcal{D}$. Through this cost we are able to ensure that the weight distribution learns a rich representation and also does not overfit. The Gaussian perturbations in each mini-batch allows the gradient estimates of the cost to be unbiased. In our sequence encoder, we obtain ReLU transformation of the final embeddings $h$ in a final linear layer. The weight distribution in the linear layer are learned using \textit{Bayes by Backprop}. We also tried other model layers for learning parameter distribution (given in \cite{sharma2023probabilistic}).

\subsection{Uncertainty Based Learning (UBL)}

It is also imperative to manage uncertainty in complex deep learning architectures that may arise due to imperfect data. This can be achieved by penalizing static characteristics estimates with higher uncertainty. Uncertainty estimates from probabilistic models can therefore enable formulation of a regularization scheme to obtain lower uncertainty estimates. We can penalize the pseudo-inverse loss (Equation~\ref{eq:Linv}) such that the characteristics with higher uncertainty in the estimates will have higher loss due to bigger penalty coefficients $w$, such that $w \propto ||\hat{Z} - \bar{Z}||$. In our work, we prove that the optimal coefficients that penalize the pseudoinverse loss the most is the eigenvector that corresponds with the largest eigenvalue of $\sigma$, the epistemic uncertainty matrix \cite{sharma2023probabilistic}.

\section{Results}
\textbf{Dataset:} We use the CAMELS dataset, which is a publicly available hydrology dataset for multiple hydrology entities (including the 531 entities that were included in our study). The input variables for the forward model are 5 time-varying weather drivers and 27 static characteristics about the entities (climate features, soil-based, and geo-morphology-based features are included in the study. These affect the streamflow). The response variable is streamflow values. In practical setting, static characteristics information for all the entities may not be known. This makes it imperative to explore models that can provide inferred characteristics for predicting streamflow for all entities. In our inverse model, the streamflow - weather time series are used for learning static characteristics.

\textbf{Experimental Setup:} Daily data from years 1980 - 2000 are used for training, years 2000 - 2005 are used for validation, and year 2005 - 2015 are used for testing. We stride over half a year and use a year as lookback period for making predictions. We divide the river basins and put 400 of the river basins into train and the rest 131 into a test set.  We report NSE (Nash-Sutcliff Efficiency is a measure similar to $R^2$ and is used to measure prediction performance in time-series hydrological models). To evaluate the uncertainty estimates, we report dispersion, coverage rate, and prediction interval width. In streamflow modeling, ensemble learning has been proven to outperform individual model prediction performance \cite{kratzert2019towards, ghosh2022robust, li2022regionalization,sharma2023probabilistic}. We use an ensemble of 5 such Bayesian inverse models (BIM) to learn basin characteristic estimates and compare them with individual model predictions. More details on the experimental setup are given in the Appendix.

\begin{figure}
    \centering
    \includegraphics[scale=0.16]{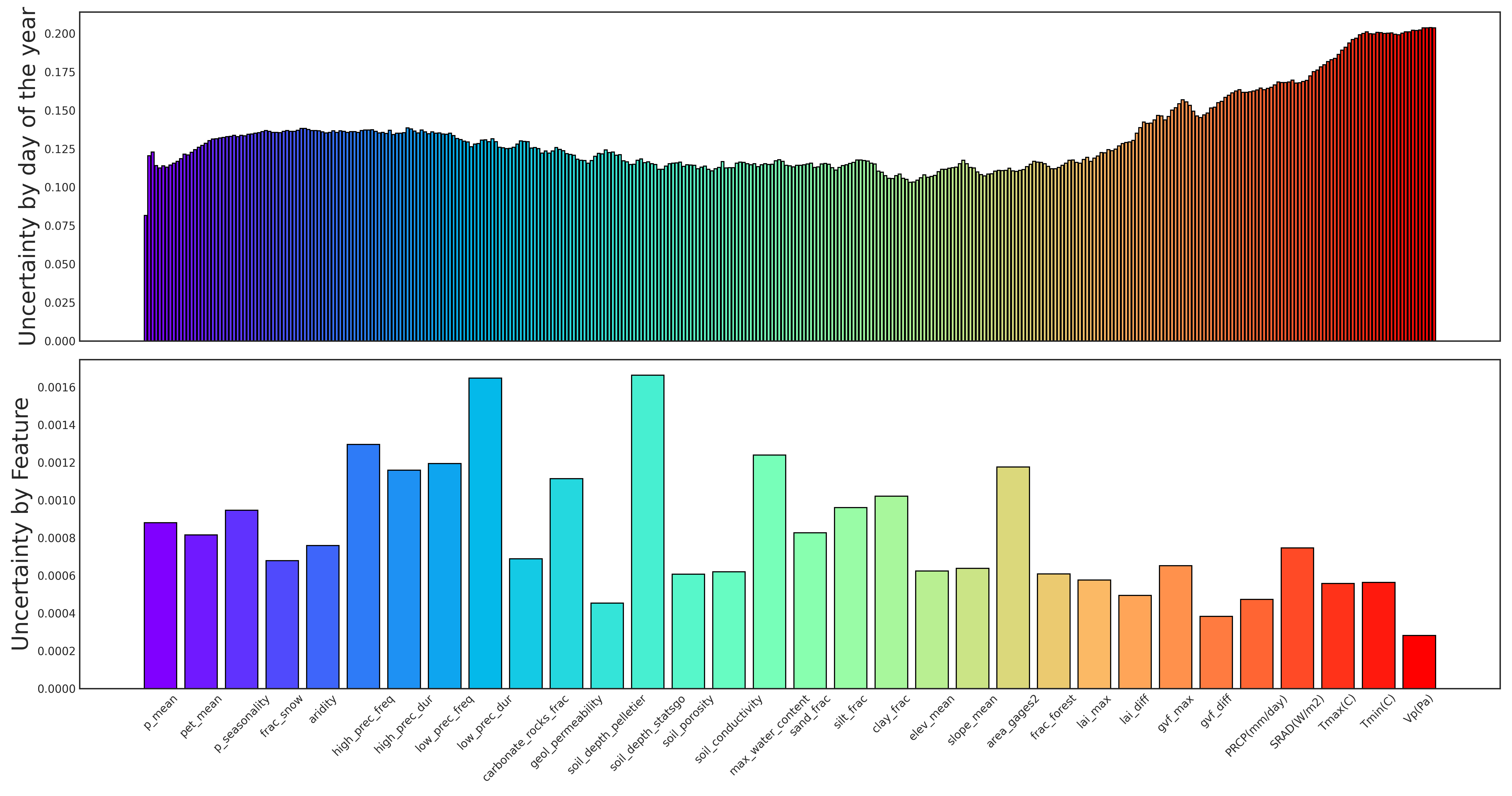}
    \caption{Model Uncertainty by day of the year and basin characteristic variables. We see higher uncertainty for periods with greater hydrological variability (precipitation and snow events). We see higher uncertainty for soil geomorphological features (turquoise/green/ orange colored middle bars) since those are the most difficult to recover from streamflow.}    
    \label{fig:model_performance}
\vspace{-10pt}
\end{figure}

\begin{table}[h]
\tiny
    \centering
    \begin{tabular}{|p{3cm}|c|p{1.2cm}|p{1.2cm}|}
    \toprule
          Model  & NSE & 63\% C.I. Coverage Rate & 95\% C.I. Coverage Rate\\
          \hline
          KGSSL & 0.6556 & - & - \\
          
          BIM  & \textbf{0.6858} & 0.8169 & 0.9386 \\
          KGSSL (UBL) & 0.6587 & - & -  \\
          BIM (UBL) & 0.6669 & \textbf{0.8220} & \textbf{0.9783}\\
          \bottomrule
    \end{tabular}
    \caption{\small Static reconstruction NSE and coverage rate. We can compare the static characteristic reconstruction NSE values among the deterministic and probabilistic models. Probabilistic models also ensure that our predictions will lie within the (mean $\pm z_{\alpha}$  s.d) interval.}
    \label{tab:inverse}
\vspace{-15pt}
\end{table}

\subsection{Static Characteristic Estimation}

The Bayesian inverse model can be compared with the state-of-the-art static characteristic estimation model, KGSSL \cite{ghosh2022robust}, in terms of prediction NSE. Table~\ref{tab:inverse} shows these results along with UBL variants, wherein, we also learn penalty coefficients for different static variables penalizing those predictions that have higher epistemic uncertainty. Since KGSSL is a deterministic model, it is unable to provide coverage rate. BIM achieves higher NSE and the calibration of uncertainty using UBL mitigates the problem of under-coverage for the 95\% confidence interval coverage rate. In our work, we have also shown that the UBL based calibration of uncertainty results in reducing the temporal artifacts in static characteristic predictions by 17\% and also reduces the epistemic uncertainty by 36\% \cite{sharma2023probabilistic}.

\begin{table}[h]
    \tiny
    \centering
    \begin{tabular}{|p{3.5cm}|c|c|}
    \toprule
          Forward Model Input  & Average NSE & Ensemble NSE\\
          \hline
          Baseline LSTM (original static characteristics)  & 0.7031 & 0.7238 \\
          KGSSL Estimates & 0.7501 & 0.7574 \\
          IM + CD Estimates & 0.7308  & 0.7502 \\
          IM + VAE Estimates & 0.7333 & 0.7565 \\
          BIM Estimates  & 0.7561 & 0.7597 \\
          KGSSL (UBL) Estimates & 0.7611 & 0.7582 \\
          BIM (UBL) Estimates & \textbf{0.7636} & \textbf{0.7659}\\
          \bottomrule
    \end{tabular}
    \caption{\small NSE in forward model streamflow prediction using reconstructed static characteristics as input. Over 5 runs, we build 5 inverse and forward models. Average NSE is average of test NSEs obtained from each forward model. Ensemble NSE is computed from average of predictions from the 5 runs. IM stands for Inverse Model and BIM stands for Bayesian Inverse Model.}
    \label{tab:forward}
\vspace{-25pt}
\end{table}

\begin{figure*}[h]
\centering
\begin{tabular}{c}
 \subcaptionbox{\label{fig:pred1} Test Set Prediction. Streamflow in log scale.   \vspace{-2pt}}{\includegraphics[scale=0.36]{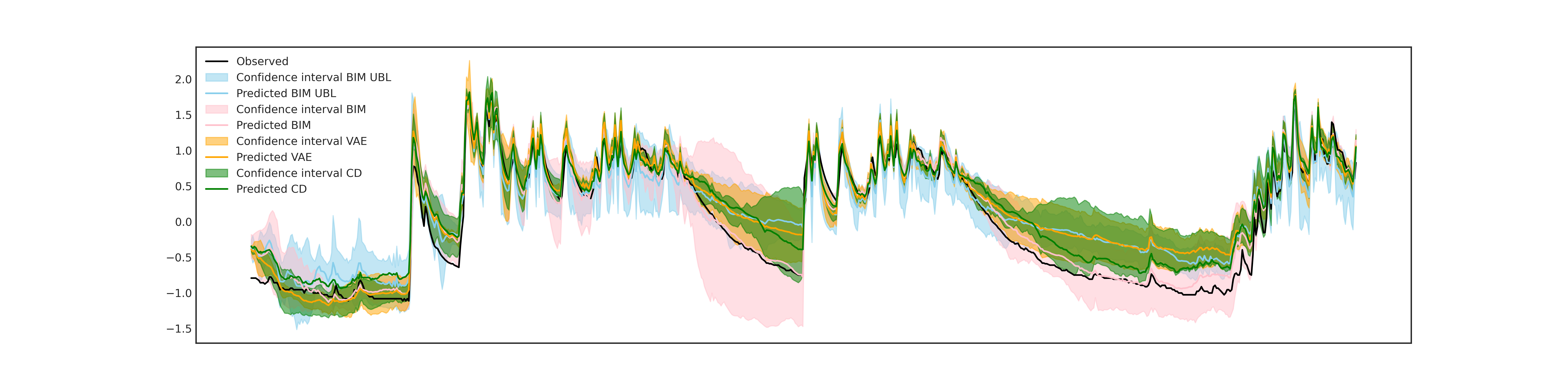}} \\
 
 \subcaptionbox{\label{fig:pred2} Test Set Prediction. Streamflow in log scale.   \vspace{-5pt}}{\includegraphics[scale=0.36]{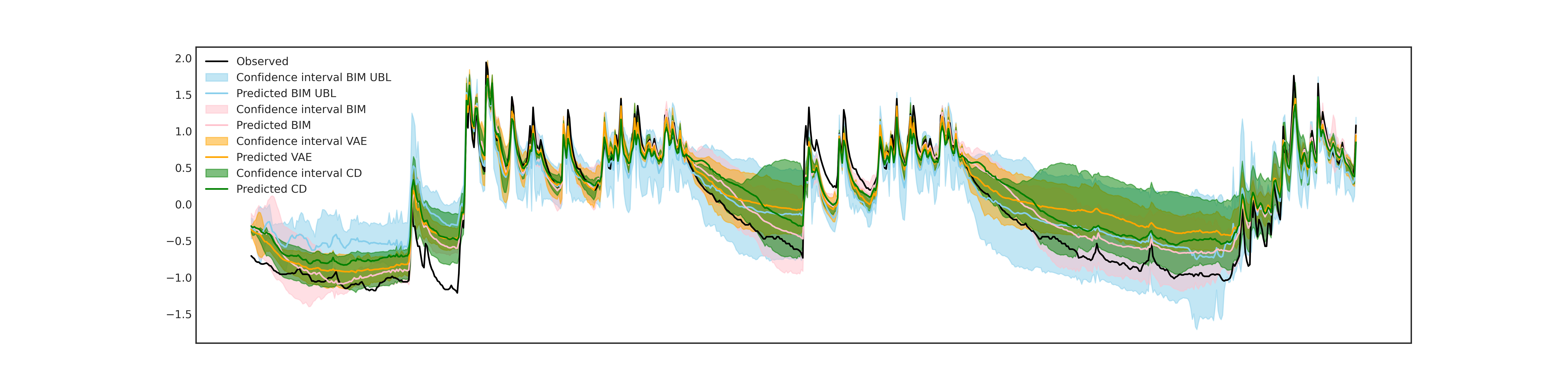}} \\ 

\end{tabular}
\caption{Random Test Sample Predictions. Pink line is proposed method and black line is ground truth streamflow.}
\label{fig:predictions}
\vspace{-10pt}
\end{figure*}

\subsection{Streamflow estimation}

We can also evaluate the utility of static characteristics predictions based on how they impact streamflow estimation in the forward model setup. We use LSTM model as the forward model since it has been proven to be state-of-the-art in streamflow prediction \cite{kratzert2019towards}. We evaluate how the forward model performance changes when we use estimates of static characteristics as input instead of the observed values. Table~\ref{tab:forward} showcases the forward model performance results, comparing using original static characteristics as input (first row in the table) to the LSTM as opposed to using predictions from the inverse model as input to forward model for forecasting streamflow in test set years. The BIM provides the best streamflow prediction model performance for individual model results (average NSE column) and ensemble results (Ensemble NSE column).



 



\begin{figure}
    \centering
    \includegraphics[scale=0.4]{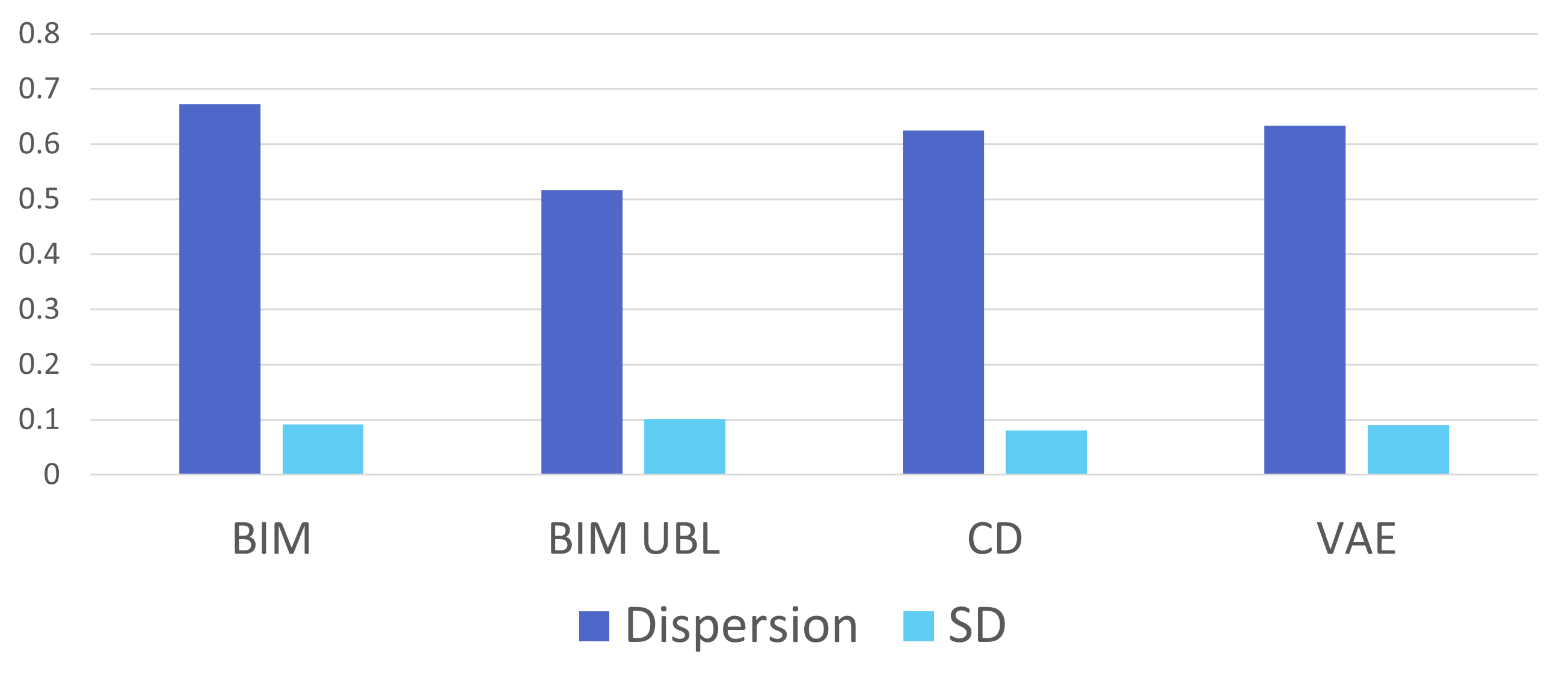}
    \caption{Uncertainty Statistics - Dispersion and standard deviation (SD) in epistemic uncertainty in inverse model}    
    \label{fig:dispersion}
\vspace{-10pt}
\end{figure}

\begin{figure}
    \centering
    \includegraphics[scale=0.36]{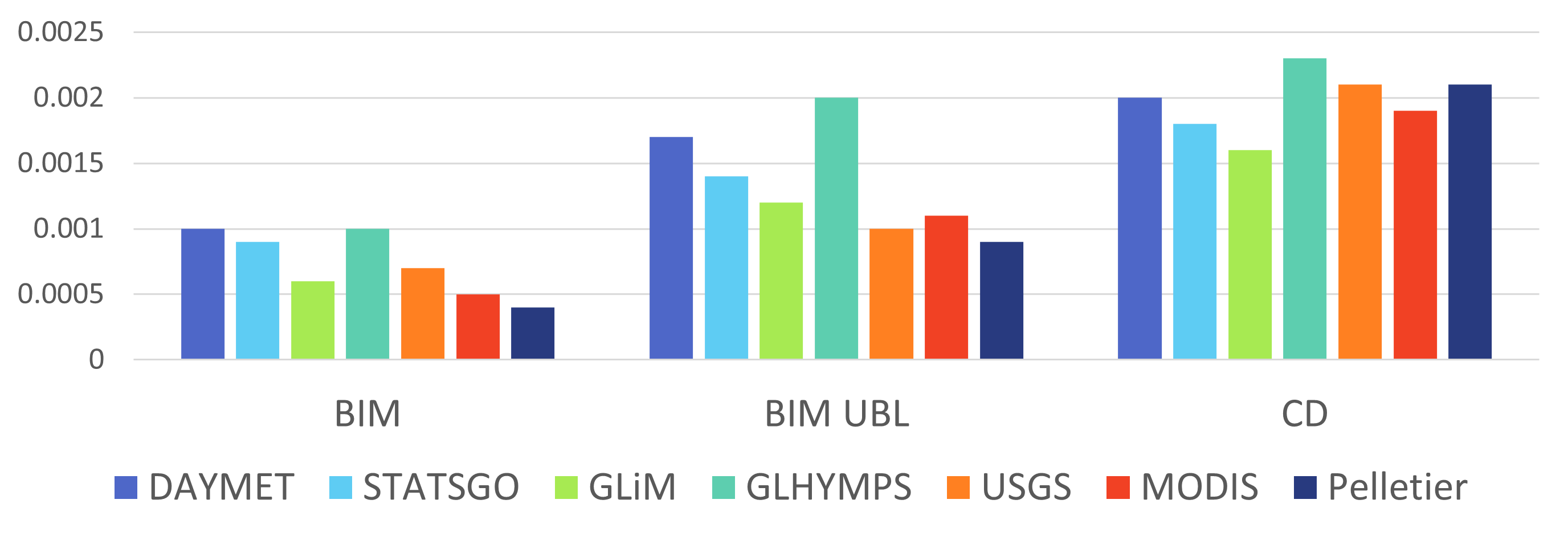}
    \caption{Uncertainty in variables by data sources used to derive the CAMELS data variables}    
    \label{fig:datasource}
\vspace{-15pt}
\end{figure}

\begin{figure}[h]
\centering
\begin{tabular}{cc}

 \subcaptionbox{\label{fig:uncdatatype} Uncertainty in input reconstructions and output of the inverse model }{\includegraphics[scale=0.3]{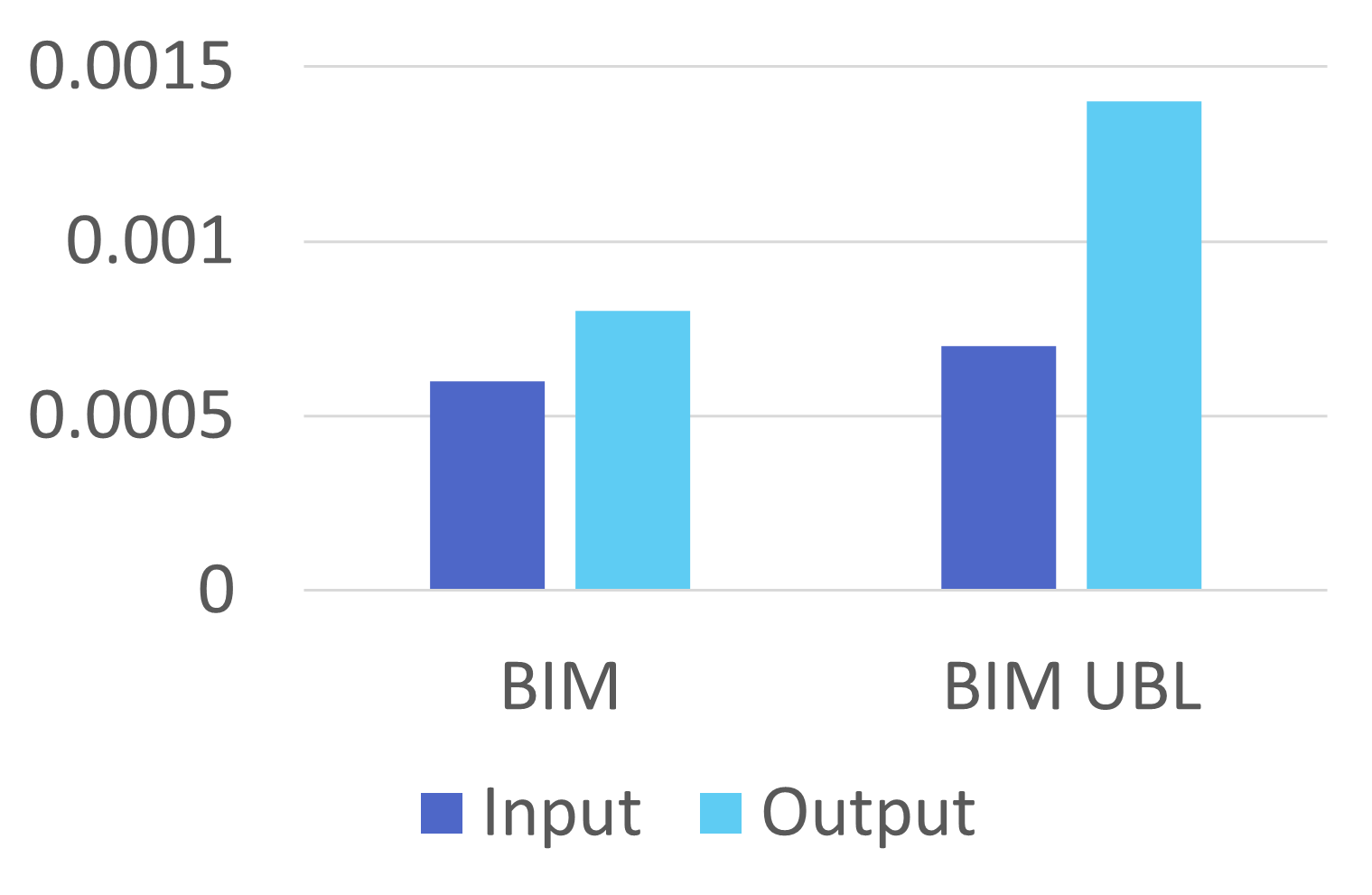}} &
 
 \subcaptionbox{\label{fig:uncphysical} Uncertainty in predictions within and outside the observed range  }{\includegraphics[scale=0.3]{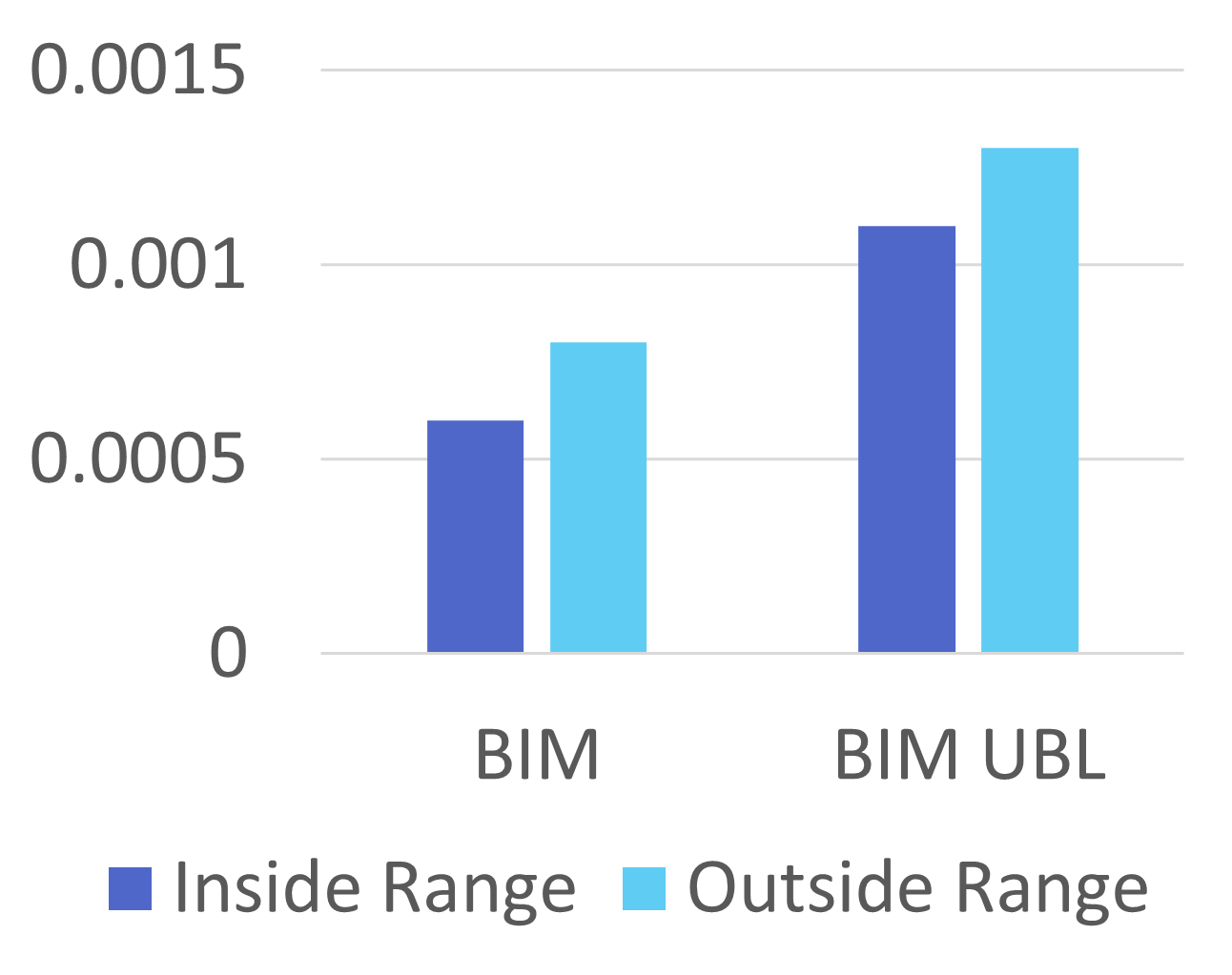}} \\ 

\end{tabular}
\caption{Uncertainty Analysis}
\label{fig:unc_breakdowns}

\vspace{-15pt}
\end{figure}


\begin{figure*}[h]

\centering
\begin{tabular}{ccc}

 \subcaptionbox{\label{fig:train-test-map} Train Test Split by Location }{\includegraphics[scale=0.15]{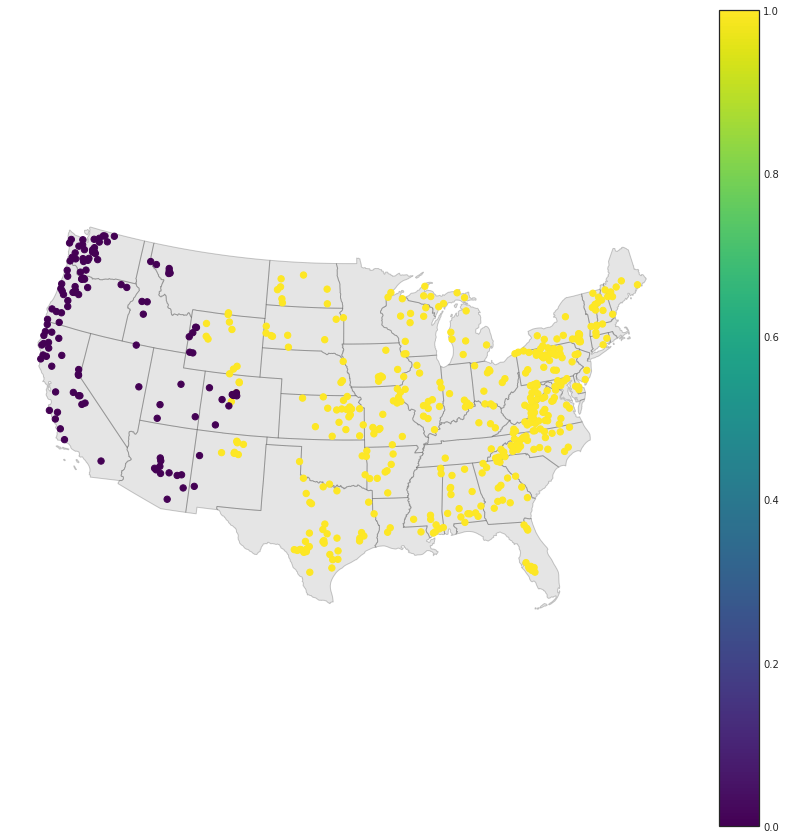}} &
 \subcaptionbox{\label{fig:unc} BIM Uncertainty in Streamflow }{\includegraphics[scale=0.15]{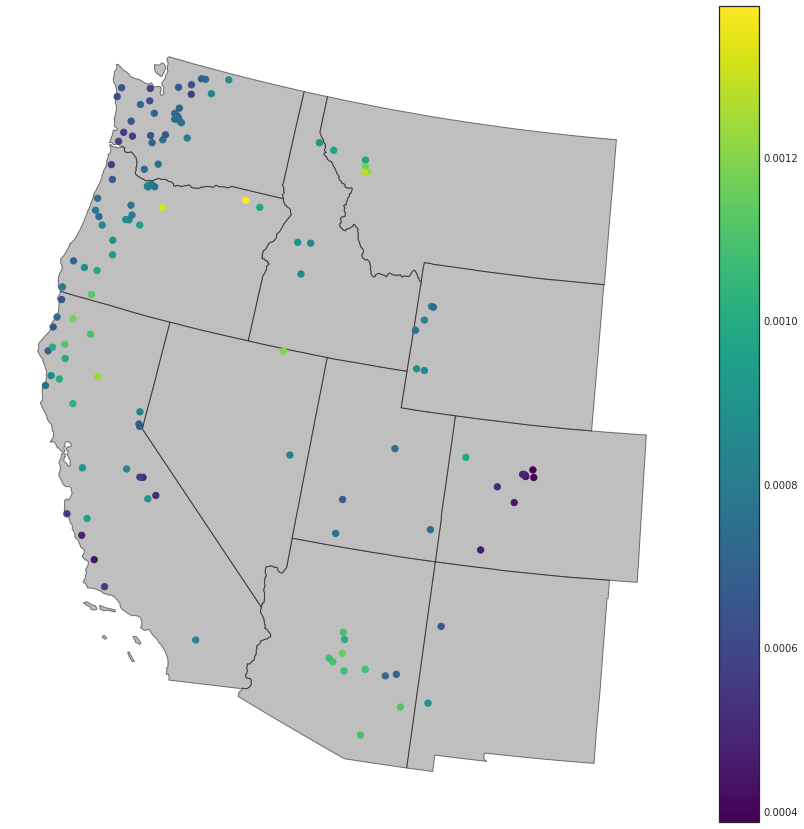}} &
 
 \subcaptionbox{\label{fig:nse-map} BIM NSE }{\includegraphics[scale=0.15]{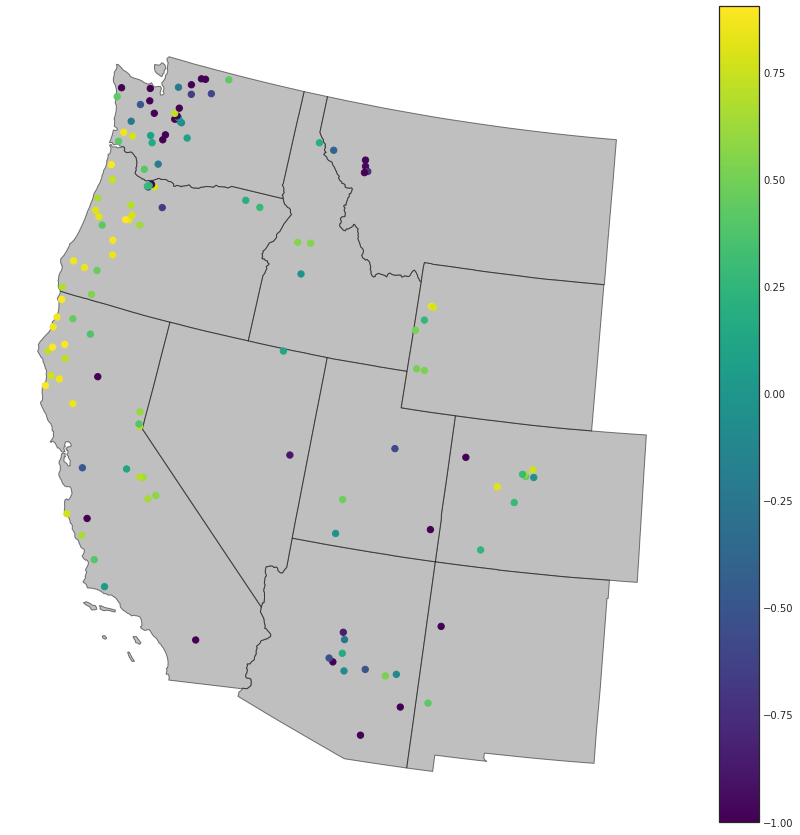}} \\ 

\end{tabular}
\caption{BIM Results by Test Locations}
\label{fig:map}

\vspace{-10pt}
\end{figure*}

\subsection{Uncertainty Quantification}

To evaluate the quality of uncertainty estimates, we compare the Bayesian inverse modeling framework against two other popular uncertainty quantification methods (outlined in Figure~\ref{fig:bim}). The first variant (IM + VAE) uses the reparameterization technique \cite{kingma2015variational} to estimate posterior distribution for the hidden encodings as part of a variational autoencoder framework. The second variant (IM + CD) uses concrete dropout method to learn the dropout rate in a linear layer that enables estimation of the posterior for the hidden encodings \cite{gal2017concrete}.

Inverse model performance for IM+VAE and IM+CD models are significantly lower than other methods. The basin characteristic estimates are still useful in their ability to leverage spatial heterogeneity in multi-basin streamflow modeling. This is in agreement with previous literature suggesting that randomly generated vectors or noisy characteristics can still enable LSTM to learn hydrological behavior and sustain benchmark streamflow prediction performance \cite{li2022regionalization}. While, Table~\ref{tab:forward} compares the streamflow prediction capacity of these frameworks, in Figure~\ref{fig:predictions}, we compare the uncertainty estimates in individual predictions for randomly selected samples in the test set. We can see the BIM predictions (pink line) are relatively closer to the ground truth streamflow (black line). For other baselines, when the predictions are far from ground truth, even the 95\% confidence intervals are unable to capture the ground truth streamflow values. The higher streamflow values relate to precipitation events while the slow decline after that relates to recession baseflow. The larger confidence bounds during the recession baseflow period suggests that predictions are more uncertain in these periods where additional water beyond direct precipitation is impacting streamflow. This may be because of soil-based heterogeneity in baseflow, which is difficult to estimate.

We can also compare the statistical consistency of uncertainty estimates as part of Figure~\ref{fig:dispersion}. While a similar root mean squared calibration error of around 7.04 for all methods suggests similarity in statistical consistency, we can further compare the dispersion in the uncertainty distribution to evaluate the quality of uncertainty estimates. Higher dispersion for BIM uncertainty estimates suggests that it is a more disperse model that can potentially be more robust to distributional shifts during model inference \cite{psaros2023uncertainty}. Lower standard deviation values for uncertainty estimates also suggest better prediction interval width and improved sharpness. However, concrete dropout achieves a sharper uncertainty distribution compared to BIM.

Since the ground truth static variables are obtained from different data sources in the CAMELS dataset, in Figure~\ref{fig:datasource}, we evaluate the uncertainty associated with any noise in these different data sources. Within the methods used for uncertainty quantification, VAE has the highest uncertainty. DAYMET, STATSGO and GLHYMPS has relatively higher uncertainty estimates, while GLim, MODIS and Pelletier result in slightly lower uncertainties. These results are in agreement with the evidence of uncertainty in these datasets suggested in previous literature \cite{Addor2017}.

The better statistical consistency of BIM methods suggests that these uncertainty estimates are relatively more trustworthy. These uncertainty estimates can potentially be used to derive further insights. For instance, uncertainty can be compared between outputs that have different levels of supervision. Input reconstructions obtained in the inverse model decoder have a better source of supervision, while the static regressor outputs have a lower level of supervision. In Sub-figure~\ref{fig:uncdatatype}, we can see that the uncertainties associated with the input reconstructions are lower than the static regressor output uncertainties. This is also reflected in Figure~\ref{fig:model_performance}. Similarly, predictions that lie outside the observed range should have higher uncertainty due to their implausibility. This is reflected in Sub-figure~\ref{fig:uncphysical}, where we see a difference in uncertainty in predictions that lie within and outside the observed value ranges in the inverse model. 

\vspace{-10pt}

\section{Discussion}

In ML applications, where prediction models may be used by stakeholders for operational decision-making, integrating uncertainty quantification methods improve the model's utility and explainability. In hydrology, a probabilistic inverse model offers us the ability to infer basin characteristics that are more trustworthy. This eliminates the need for thorough curating of large datasets that might be very expensive and time-consuming \cite{gebru2017fine}. In our framework, we quantify uncertainty arising from different sources. Once such classification can be by different data sources that were used to create the CAMELS data. For instance, uncertainty is higher in static variable predictions from STATSGO data. This may be because of bias in estimates for soil depth and soil-related features that were used to create the CAMELS dataset. This is also suggested in the CAMELS data paper \cite{Addor2017}. Similarly, uncertainty estimates may also shed light on how the model behaves for different time periods (Figure~\ref{fig:model_performance}) and under different dominant hydrological processes (Figure~\ref{fig:predictions}). 

Uncertainty estimates can also offer insights into spatial variability in hydrological processes over different river basins. For instance, Figure~\ref{fig:map} presents the BIM uncertainty estimates and NSE scores for the test set locations. We can notice higher uncertainties over the pacific northwest region that may arise from a higher frequency of precipitation events. Moreover, we see also see higher NSE for river basins in Northern California and Oregon as compared to other basins. There are differences in the dominant hydrological processes in the river basins that may have resulted in differences in model performance. For instance, Washington gets more high flow and high precipitation days as compared to Oregon and California. Washington also has a higher baseflow index and also higher soil depth. 

In hydrology, probabilistic inverse modeling can offer many insights. Better reconstructions for variables like soil porosity and conductivity imply their impact on the streamflow generation process is easily predictable as they govern soil water storage and permeability behavior more closely. In contrast, variables like carbonate rock fraction are poorer because the fraction by itself is not directly related to flow characteristics; a more predictable alternate would be the fraction of solution channels. This effect is also showcased in the lower prediction skills of the inverse model and higher uncertainty. Therefore, model users can be more cautious about inferred basin characteristics that have higher uncertainty.

Quantifying uncertainty in hydrology can aid in assessing the reliability of the models, establishing decision thresholds for acceptable levels of uncertainty, and identification of high-risk scenarios - all of which can enable improved explainability of ML models and can provide decision-makers with a clear understanding of when to trust the model's outputs.

\section{Acknowledgements}
This work was funded by the NSF award 2313174 and 2134904. Access to computing facilities was provided by the Minnesota Supercomputing Institute.

\clearpage
\bibliographystyle{ACM-Reference-Format}
\bibliography{main}


\begin{thebibliography}{37}


\ifx \showCODEN    \undefined \def \showCODEN     #1{\unskip}     \fi
\ifx \showDOI      \undefined \def \showDOI       #1{#1}\fi
\ifx \showISBNx    \undefined \def \showISBNx     #1{\unskip}     \fi
\ifx \showISBNxiii \undefined \def \showISBNxiii  #1{\unskip}     \fi
\ifx \showISSN     \undefined \def \showISSN      #1{\unskip}     \fi
\ifx \showLCCN     \undefined \def \showLCCN      #1{\unskip}     \fi
\ifx \shownote     \undefined \def \shownote      #1{#1}          \fi
\ifx \showarticletitle \undefined \def \showarticletitle #1{#1}   \fi
\ifx \showURL      \undefined \def \showURL       {\relax}        \fi
\providecommand\bibfield[2]{#2}
\providecommand\bibinfo[2]{#2}
\providecommand\natexlab[1]{#1}
\providecommand\showeprint[2][]{arXiv:#2}

\bibitem[Addor et~al\mbox{.}(2017)]%
        {Addor2017}
\bibfield{author}{\bibinfo{person}{Nans Addor} {et~al\mbox{.}}}
  \bibinfo{year}{2017}\natexlab{}.
\newblock \showarticletitle{{The CAMELS data set: Catchment attributes and
  meteorology for large-sample studies}}.
\newblock \bibinfo{journal}{\emph{Hydrology and Earth System Sciences}}
  \bibinfo{volume}{21}, \bibinfo{number}{10} (\bibinfo{year}{2017}),
  \bibinfo{pages}{5293--5313}.
\newblock
\showISSN{16077938}
\urldef\tempurl%
\url{https://doi.org/10.5194/hess-21-5293-2017}
\showDOI{\tempurl}


\bibitem[Asim et~al\mbox{.}(2020)]%
        {asim2020invertible}
\bibfield{author}{\bibinfo{person}{Muhammad Asim} {et~al\mbox{.}}}
  \bibinfo{year}{2020}\natexlab{}.
\newblock \showarticletitle{Invertible generative models for inverse problems:
  mitigating representation error and dataset bias}. In
  \bibinfo{booktitle}{\emph{International Conference on Machine Learning}}.
  PMLR, \bibinfo{pages}{399--409}.
\newblock


\bibitem[Blundell et~al\mbox{.}(2015)]%
        {blundell2015weight}
\bibfield{author}{\bibinfo{person}{Charles Blundell}, \bibinfo{person}{Julien
  Cornebise}, \bibinfo{person}{Koray Kavukcuoglu}, {and} \bibinfo{person}{Daan
  Wierstra}.} \bibinfo{year}{2015}\natexlab{}.
\newblock \showarticletitle{Weight uncertainty in neural network}. In
  \bibinfo{booktitle}{\emph{International Conference on Machine Learning}}.
  PMLR, \bibinfo{pages}{1613--1622}.
\newblock


\bibitem[Carbone et~al\mbox{.}(2020)]%
        {carbone2020robustness}
\bibfield{author}{\bibinfo{person}{Ginevra Carbone}, \bibinfo{person}{Matthew
  Wicker}, \bibinfo{person}{Luca Laurenti}, \bibinfo{person}{Andrea Patane},
  \bibinfo{person}{Luca Bortolussi}, {and} \bibinfo{person}{Guido
  Sanguinetti}.} \bibinfo{year}{2020}\natexlab{}.
\newblock \showarticletitle{Robustness of bayesian neural networks to
  gradient-based attacks}.
\newblock \bibinfo{journal}{\emph{Advances in Neural Information Processing
  Systems}}  \bibinfo{volume}{33} (\bibinfo{year}{2020}),
  \bibinfo{pages}{15602--15613}.
\newblock


\bibitem[Cardelli et~al\mbox{.}(2019)]%
        {cardelli2019statistical}
\bibfield{author}{\bibinfo{person}{Luca Cardelli}, \bibinfo{person}{Marta
  Kwiatkowska}, \bibinfo{person}{Luca Laurenti}, \bibinfo{person}{Nicola
  Paoletti}, \bibinfo{person}{Andrea Patane}, {and} \bibinfo{person}{Matthew
  Wicker}.} \bibinfo{year}{2019}\natexlab{}.
\newblock \showarticletitle{Statistical guarantees for the robustness of
  Bayesian neural networks}.
\newblock \bibinfo{journal}{\emph{arXiv preprint arXiv:1903.01980}}
  (\bibinfo{year}{2019}).
\newblock


\bibitem[Dao et~al\mbox{.}(2021)]%
        {dao2021improving}
\bibfield{author}{\bibinfo{person}{Phuong~D Dao} {et~al\mbox{.}}}
  \bibinfo{year}{2021}\natexlab{}.
\newblock \showarticletitle{Improving hyperspectral image segmentation by
  applying inverse noise weighting and outlier removal for optimal scale
  selection}.
\newblock \bibinfo{journal}{\emph{ISPRS Journal of Photogrammetry and Remote
  Sensing}}  \bibinfo{volume}{171} (\bibinfo{year}{2021}),
  \bibinfo{pages}{348--366}.
\newblock


\bibitem[Daw et~al\mbox{.}(2021)]%
        {daw2021pid}
\bibfield{author}{\bibinfo{person}{Arka Daw}, \bibinfo{person}{M Maruf}, {and}
  \bibinfo{person}{Anuj Karpatne}.} \bibinfo{year}{2021}\natexlab{}.
\newblock \showarticletitle{PID-GAN: A GAN Framework based on a
  Physics-informed Discriminator for Uncertainty Quantification with Physics}.
  In \bibinfo{booktitle}{\emph{Proceedings of the 27th ACM SIGKDD Conference on
  Knowledge Discovery \& Data Mining}}. \bibinfo{pages}{237--247}.
\newblock


\bibitem[Friston et~al\mbox{.}(2007)]%
        {friston2007variational}
\bibfield{author}{\bibinfo{person}{Karl Friston},
  \bibinfo{person}{J{\'e}r{\'e}mie Mattout}, \bibinfo{person}{Nelson
  Trujillo-Barreto}, \bibinfo{person}{John Ashburner}, {and}
  \bibinfo{person}{Will Penny}.} \bibinfo{year}{2007}\natexlab{}.
\newblock \showarticletitle{Variational free energy and the Laplace
  approximation}.
\newblock \bibinfo{journal}{\emph{Neuroimage}} \bibinfo{volume}{34},
  \bibinfo{number}{1} (\bibinfo{year}{2007}), \bibinfo{pages}{220--234}.
\newblock


\bibitem[Gal et~al\mbox{.}(2017)]%
        {gal2017concrete}
\bibfield{author}{\bibinfo{person}{Yarin Gal}, \bibinfo{person}{Jiri Hron},
  {and} \bibinfo{person}{Alex Kendall}.} \bibinfo{year}{2017}\natexlab{}.
\newblock \showarticletitle{Concrete dropout}.
\newblock \bibinfo{journal}{\emph{arXiv preprint arXiv:1705.07832}}
  (\bibinfo{year}{2017}).
\newblock


\bibitem[Gebru et~al\mbox{.}(2017)]%
        {gebru2017fine}
\bibfield{author}{\bibinfo{person}{Timnit Gebru}, \bibinfo{person}{Judy
  Hoffman}, {and} \bibinfo{person}{Li Fei-Fei}.}
  \bibinfo{year}{2017}\natexlab{}.
\newblock \showarticletitle{Fine-grained recognition in the wild: A multi-task
  domain adaptation approach}. In \bibinfo{booktitle}{\emph{Proceedings of the
  IEEE international conference on computer vision}}.
  \bibinfo{pages}{1349--1358}.
\newblock


\bibitem[Ghimire et~al\mbox{.}(2021)]%
        {ghimire2021streamflow}
\bibfield{author}{\bibinfo{person}{Sujan Ghimire},
  \bibinfo{person}{Zaher~Mundher Yaseen}, \bibinfo{person}{Aitazaz~A Farooque},
  \bibinfo{person}{Ravinesh~C Deo}, \bibinfo{person}{Ji Zhang}, {and}
  \bibinfo{person}{Xiaohui Tao}.} \bibinfo{year}{2021}\natexlab{}.
\newblock \showarticletitle{Streamflow prediction using an integrated
  methodology based on convolutional neural network and long short-term memory
  networks}.
\newblock \bibinfo{journal}{\emph{Scientific Reports}} \bibinfo{volume}{11},
  \bibinfo{number}{1} (\bibinfo{year}{2021}), \bibinfo{pages}{1--26}.
\newblock


\bibitem[Ghosh et~al\mbox{.}(2022)]%
        {ghosh2022robust}
\bibfield{author}{\bibinfo{person}{Rahul Ghosh}, \bibinfo{person}{Arvind
  Renganathan}, \bibinfo{person}{Kshitij Tayal}, \bibinfo{person}{Xiang Li},
  \bibinfo{person}{Ankush Khandelwal}, \bibinfo{person}{Xiaowei Jia},
  \bibinfo{person}{Christopher Duffy}, \bibinfo{person}{John Nieber}, {and}
  \bibinfo{person}{Vipin Kumar}.} \bibinfo{year}{2022}\natexlab{}.
\newblock \showarticletitle{Robust Inverse Framework using Knowledge-guided
  Self-Supervised Learning: An application to Hydrology}. In
  \bibinfo{booktitle}{\emph{Proceedings of the 28th ACM SIGKDD Conference on
  Knowledge Discovery and Data Mining}}. \bibinfo{pages}{465--474}.
\newblock


\bibitem[Goodfellow et~al\mbox{.}(2013)]%
        {goodfellow2013maxout}
\bibfield{author}{\bibinfo{person}{Ian Goodfellow}, \bibinfo{person}{David
  Warde-Farley}, \bibinfo{person}{Mehdi Mirza}, \bibinfo{person}{Aaron
  Courville}, {and} \bibinfo{person}{Yoshua Bengio}.}
  \bibinfo{year}{2013}\natexlab{}.
\newblock \showarticletitle{Maxout networks}. In
  \bibinfo{booktitle}{\emph{International conference on machine learning}}.
  PMLR, \bibinfo{pages}{1319--1327}.
\newblock


\bibitem[Graves(2011)]%
        {graves2011practical}
\bibfield{author}{\bibinfo{person}{Alex Graves}.}
  \bibinfo{year}{2011}\natexlab{}.
\newblock \showarticletitle{Practical variational inference for neural
  networks}.
\newblock \bibinfo{journal}{\emph{Advances in neural information processing
  systems}}  \bibinfo{volume}{24} (\bibinfo{year}{2011}),
  \bibinfo{pages}{2348--2356}.
\newblock


\bibitem[Hanson and Pratt(1988)]%
        {hanson1988comparing}
\bibfield{author}{\bibinfo{person}{Stephen Hanson} {and}
  \bibinfo{person}{Lorien Pratt}.} \bibinfo{year}{1988}\natexlab{}.
\newblock \showarticletitle{Comparing biases for minimal network construction
  with back-propagation}.
\newblock \bibinfo{journal}{\emph{Advances in neural information processing
  systems}}  \bibinfo{volume}{1} (\bibinfo{year}{1988}),
  \bibinfo{pages}{177--185}.
\newblock


\bibitem[Her et~al\mbox{.}(2019)]%
        {her2019uncertainty}
\bibfield{author}{\bibinfo{person}{Younggu Her}, \bibinfo{person}{Seung-Hwan
  Yoo}, \bibinfo{person}{Jaepil Cho}, \bibinfo{person}{Syewoon Hwang},
  \bibinfo{person}{Jaehak Jeong}, {and} \bibinfo{person}{Chounghyun Seong}.}
  \bibinfo{year}{2019}\natexlab{}.
\newblock \showarticletitle{Uncertainty in hydrological analysis of climate
  change: multi-parameter vs. multi-GCM ensemble predictions}.
\newblock \bibinfo{journal}{\emph{Scientific reports}} \bibinfo{volume}{9},
  \bibinfo{number}{1} (\bibinfo{year}{2019}), \bibinfo{pages}{1--22}.
\newblock


\bibitem[Jaakkola and Jordan(2000)]%
        {jaakkola2000bayesian}
\bibfield{author}{\bibinfo{person}{Tommi~S Jaakkola} {and}
  \bibinfo{person}{Michael~I Jordan}.} \bibinfo{year}{2000}\natexlab{}.
\newblock \showarticletitle{Bayesian parameter estimation via variational
  methods}.
\newblock \bibinfo{journal}{\emph{Statistics and Computing}}
  \bibinfo{volume}{10}, \bibinfo{number}{1} (\bibinfo{year}{2000}),
  \bibinfo{pages}{25--37}.
\newblock


\bibitem[Kang et~al\mbox{.}(2016)]%
        {kang2016shakeout}
\bibfield{author}{\bibinfo{person}{Guoliang Kang}, \bibinfo{person}{Jun Li},
  {and} \bibinfo{person}{Dacheng Tao}.} \bibinfo{year}{2016}\natexlab{}.
\newblock \showarticletitle{Shakeout: A new regularized deep neural network
  training scheme}. In \bibinfo{booktitle}{\emph{Thirtieth AAAI Conference on
  Artificial Intelligence}}.
\newblock


\bibitem[Kingma et~al\mbox{.}(2015)]%
        {kingma2015variational}
\bibfield{author}{\bibinfo{person}{Durk~P Kingma}, \bibinfo{person}{Tim
  Salimans}, {and} \bibinfo{person}{Max Welling}.}
  \bibinfo{year}{2015}\natexlab{}.
\newblock \showarticletitle{Variational dropout and the local
  reparameterization trick}.
\newblock \bibinfo{journal}{\emph{Advances in neural information processing
  systems}}  \bibinfo{volume}{28} (\bibinfo{year}{2015}),
  \bibinfo{pages}{2575--2583}.
\newblock


\bibitem[Kratzert et~al\mbox{.}(2019)]%
        {kratzert2019towards}
\bibfield{author}{\bibinfo{person}{Frederik Kratzert} {et~al\mbox{.}}}
  \bibinfo{year}{2019}\natexlab{}.
\newblock \showarticletitle{Towards learning universal, regional, and local
  hydrological behaviors via machine learning applied to large-sample
  datasets}.
\newblock \bibinfo{journal}{\emph{Hydrology and Earth System Sciences}}
  \bibinfo{volume}{23}, \bibinfo{number}{12} (\bibinfo{year}{2019}),
  \bibinfo{pages}{5089--5110}.
\newblock


\bibitem[Lavin et~al\mbox{.}(2021)]%
        {lavin2021simulation}
\bibfield{author}{\bibinfo{person}{Alexander Lavin}, \bibinfo{person}{Hector
  Zenil}, \bibinfo{person}{Brooks Paige}, \bibinfo{person}{David Krakauer},
  \bibinfo{person}{Justin Gottschlich}, \bibinfo{person}{Tim Mattson},
  \bibinfo{person}{Anima Anandkumar}, \bibinfo{person}{Sanjay Choudry},
  \bibinfo{person}{Kamil Rocki}, \bibinfo{person}{At{\i}l{\i}m~G{\"u}ne{\c{s}}
  Baydin}, {et~al\mbox{.}}} \bibinfo{year}{2021}\natexlab{}.
\newblock \showarticletitle{Simulation Intelligence: Towards a New Generation
  of Scientific Methods}.
\newblock \bibinfo{journal}{\emph{arXiv preprint arXiv:2112.03235}}
  (\bibinfo{year}{2021}).
\newblock


\bibitem[Li et~al\mbox{.}(2022)]%
        {li2022regionalization}
\bibfield{author}{\bibinfo{person}{Xiang Li}, \bibinfo{person}{Ankush
  Khandelwal}, \bibinfo{person}{Xiaowei Jia}, \bibinfo{person}{Kelly Cutler},
  \bibinfo{person}{Rahul Ghosh}, \bibinfo{person}{Arvind Renganathan},
  \bibinfo{person}{Shaoming Xu}, \bibinfo{person}{JL Nieber},
  \bibinfo{person}{Christopher~J Duffy}, \bibinfo{person}{Michael Steinbach},
  {et~al\mbox{.}}} \bibinfo{year}{2022}\natexlab{}.
\newblock \showarticletitle{Regionalization in a global hydrologic deep
  learning model: from physical descriptors to random vectors}.
\newblock  (\bibinfo{year}{2022}).
\newblock


\bibitem[Li and Liu(2016)]%
        {li2016whiteout}
\bibfield{author}{\bibinfo{person}{Yinan Li} {and} \bibinfo{person}{Fang Liu}.}
  \bibinfo{year}{2016}\natexlab{}.
\newblock \showarticletitle{Whiteout: Gaussian adaptive noise regularization in
  deep neural networks}.
\newblock \bibinfo{journal}{\emph{arXiv preprint arXiv:1612.01490}}
  (\bibinfo{year}{2016}).
\newblock


\bibitem[McMillan et~al\mbox{.}(2018)]%
        {mcmillan2018hydrological}
\bibfield{author}{\bibinfo{person}{Hilary~K McMillan}, \bibinfo{person}{Ida~K
  Westerberg}, {and} \bibinfo{person}{Tobias Krueger}.}
  \bibinfo{year}{2018}\natexlab{}.
\newblock \showarticletitle{Hydrological data uncertainty and its
  implications}.
\newblock \bibinfo{journal}{\emph{Wiley Interdisciplinary Reviews: Water}}
  \bibinfo{volume}{5}, \bibinfo{number}{6} (\bibinfo{year}{2018}),
  \bibinfo{pages}{e1319}.
\newblock


\bibitem[Neal and Hinton(1998)]%
        {neal1998view}
\bibfield{author}{\bibinfo{person}{Radford~M Neal} {and}
  \bibinfo{person}{Geoffrey~E Hinton}.} \bibinfo{year}{1998}\natexlab{}.
\newblock \showarticletitle{A view of the EM algorithm that justifies
  incremental, sparse, and other variants}.
\newblock In \bibinfo{booktitle}{\emph{Learning in graphical models}}.
  \bibinfo{publisher}{Springer}, \bibinfo{pages}{355--368}.
\newblock


\bibitem[Newman et~al\mbox{.}(2015)]%
        {newman2015gridded}
\bibfield{author}{\bibinfo{person}{Andrew~J Newman} {et~al\mbox{.}}}
  \bibinfo{year}{2015}\natexlab{}.
\newblock \showarticletitle{Gridded ensemble precipitation and temperature
  estimates for the contiguous United States}.
\newblock \bibinfo{journal}{\emph{Journal of Hydrometeorology}}
  \bibinfo{volume}{16}, \bibinfo{number}{6} (\bibinfo{year}{2015}),
  \bibinfo{pages}{2481--2500}.
\newblock


\bibitem[Pecha et~al\mbox{.}(2021)]%
        {pecha2021determination}
\bibfield{author}{\bibinfo{person}{Petr Pecha} {et~al\mbox{.}}}
  \bibinfo{year}{2021}\natexlab{}.
\newblock \showarticletitle{Determination of radiological background fields
  designated for inverse modelling during atypical low wind speed
  meteorological episode}.
\newblock \bibinfo{journal}{\emph{Atmospheric Environment}}
  \bibinfo{volume}{246} (\bibinfo{year}{2021}), \bibinfo{pages}{118105}.
\newblock


\bibitem[Psaros et~al\mbox{.}(2023)]%
        {psaros2023uncertainty}
\bibfield{author}{\bibinfo{person}{Apostolos~F Psaros}, \bibinfo{person}{Xuhui
  Meng}, \bibinfo{person}{Zongren Zou}, \bibinfo{person}{Ling Guo}, {and}
  \bibinfo{person}{George~Em Karniadakis}.} \bibinfo{year}{2023}\natexlab{}.
\newblock \showarticletitle{Uncertainty quantification in scientific machine
  learning: Methods, metrics, and comparisons}.
\newblock \bibinfo{journal}{\emph{J. Comput. Phys.}} (\bibinfo{year}{2023}),
  \bibinfo{pages}{111902}.
\newblock


\bibitem[Sharma and Chatterjee(2021)]%
        {sharma2021winsorization}
\bibfield{author}{\bibinfo{person}{Somya Sharma} {and}
  \bibinfo{person}{Snigdhansu Chatterjee}.} \bibinfo{year}{2021}\natexlab{}.
\newblock \showarticletitle{Winsorization for Robust Bayesian Neural Networks}.
\newblock \bibinfo{journal}{\emph{Entropy}} \bibinfo{volume}{23},
  \bibinfo{number}{11} (\bibinfo{year}{2021}), \bibinfo{pages}{1546}.
\newblock


\bibitem[Sharma et~al\mbox{.}(2023)]%
        {sharma2023probabilistic}
\bibfield{author}{\bibinfo{person}{Somya Sharma}, \bibinfo{person}{Rahul
  Ghosh}, \bibinfo{person}{Arvind Renganathan}, \bibinfo{person}{Xiang Li},
  \bibinfo{person}{Snigdhansu Chatterjee}, \bibinfo{person}{John Nieber},
  \bibinfo{person}{Christopher Duffy}, {and} \bibinfo{person}{Vipin Kumar}.}
  \bibinfo{year}{2023}\natexlab{}.
\newblock \showarticletitle{Probabilistic Inverse Modeling: An Application in
  Hydrology}. In \bibinfo{booktitle}{\emph{Proceedings of the 2023 SIAM
  International Conference on Data Mining (SDM)}}. SIAM,
  \bibinfo{pages}{847--855}.
\newblock


\bibitem[Srivastava et~al\mbox{.}(2014)]%
        {srivastava2014dropout}
\bibfield{author}{\bibinfo{person}{Nitish Srivastava},
  \bibinfo{person}{Geoffrey Hinton}, \bibinfo{person}{Alex Krizhevsky},
  \bibinfo{person}{Ilya Sutskever}, {and} \bibinfo{person}{Ruslan
  Salakhutdinov}.} \bibinfo{year}{2014}\natexlab{}.
\newblock \showarticletitle{Dropout: a simple way to prevent neural networks
  from overfitting}.
\newblock \bibinfo{journal}{\emph{The journal of machine learning research}}
  \bibinfo{volume}{15}, \bibinfo{number}{1} (\bibinfo{year}{2014}),
  \bibinfo{pages}{1929--1958}.
\newblock


\bibitem[Sun and Bouman(2020)]%
        {sun2020deep}
\bibfield{author}{\bibinfo{person}{He Sun} {and} \bibinfo{person}{Katherine~L
  Bouman}.} \bibinfo{year}{2020}\natexlab{}.
\newblock \showarticletitle{Deep probabilistic imaging: Uncertainty
  quantification and multi-modal solution characterization for computational
  imaging}.
\newblock \bibinfo{journal}{\emph{arXiv preprint arXiv:2010.14462}}
  \bibinfo{volume}{9} (\bibinfo{year}{2020}).
\newblock


\bibitem[Wan et~al\mbox{.}(2013)]%
        {wan2013regularization}
\bibfield{author}{\bibinfo{person}{Li Wan}, \bibinfo{person}{Matthew Zeiler},
  \bibinfo{person}{Sixin Zhang}, \bibinfo{person}{Yann Le~Cun}, {and}
  \bibinfo{person}{Rob Fergus}.} \bibinfo{year}{2013}\natexlab{}.
\newblock \showarticletitle{Regularization of neural networks using
  dropconnect}. In \bibinfo{booktitle}{\emph{International conference on
  machine learning}}. PMLR, \bibinfo{pages}{1058--1066}.
\newblock


\bibitem[Wen et~al\mbox{.}(2018)]%
        {wen2018flipout}
\bibfield{author}{\bibinfo{person}{Yeming Wen}, \bibinfo{person}{Paul Vicol},
  \bibinfo{person}{Jimmy Ba}, \bibinfo{person}{Dustin Tran}, {and}
  \bibinfo{person}{Roger Grosse}.} \bibinfo{year}{2018}\natexlab{}.
\newblock \showarticletitle{Flipout: Efficient pseudo-independent weight
  perturbations on mini-batches}.
\newblock \bibinfo{journal}{\emph{arXiv preprint arXiv:1803.04386}}
  (\bibinfo{year}{2018}).
\newblock


\bibitem[Whang et~al\mbox{.}(2021)]%
        {whang2021solving}
\bibfield{author}{\bibinfo{person}{Jay Whang}, \bibinfo{person}{Qi Lei}, {and}
  \bibinfo{person}{Alex Dimakis}.} \bibinfo{year}{2021}\natexlab{}.
\newblock \showarticletitle{Solving inverse problems with a flow-based noise
  model}. In \bibinfo{booktitle}{\emph{International Conference on Machine
  Learning}}. PMLR, \bibinfo{pages}{11146--11157}.
\newblock


\bibitem[Woolway et~al\mbox{.}(2021)]%
        {woolway2021winter}
\bibfield{author}{\bibinfo{person}{R~Iestyn Woolway} {et~al\mbox{.}}}
  \bibinfo{year}{2021}\natexlab{}.
\newblock \showarticletitle{Winter inverse lake stratification under historic
  and future climate change}.
\newblock \bibinfo{journal}{\emph{Limnology and Oceanography Letters}}
  (\bibinfo{year}{2021}).
\newblock


\bibitem[Yedidia et~al\mbox{.}(2000)]%
        {yedidia2000generalized}
\bibfield{author}{\bibinfo{person}{Jonathan~S Yedidia},
  \bibinfo{person}{William~T Freeman}, \bibinfo{person}{Yair Weiss},
  {et~al\mbox{.}}} \bibinfo{year}{2000}\natexlab{}.
\newblock \showarticletitle{Generalized belief propagation}. In
  \bibinfo{booktitle}{\emph{NIPS}}, Vol.~\bibinfo{volume}{13}.
  \bibinfo{pages}{689--695}.
\newblock


\end{thebibliography}

\clearpage
\appendix
\section{Reproducibility}
The code and data is shared \href{https://tinyurl.com/kdd23}{here}. The daily-level CAMELS dataset used in this study is available \href{https://ral.ucar.edu/solutions/products/camels}{here}. The framework is built in \href{https://pytorch.org/}{PyTorch}, with the Bayes by Backprop module being a modification from \href{https://anaconda.org/conda-forge/blitz-bayesian-pytorch}{Blitz package}. 

\section{Evaluating Model Performance:} 

To ensure there is no data leakage when using basin characteristics predictions from the inverse model to make predictions in the test set using the forward model, we estimate basin characteristics using the validation data. The inverse model, $g_{\mathcal{S}}$, is trained on training set $\mathcal{S}$ such that, $g_S: [x_t^i, y_t^i] \rightarrow z_i$. In our case, training, validation and test set are divided to test temporal generalizability. Let validation set be $\mathcal{S}_{val}$ and test set by $\mathcal{S}_{test}$. We obtain static characteristic reconstructions on validation set. Now for forward modeling, we average the validation set reconstructions over time to obtain $\hat{z}_{val}$. Since the reconstructions are supposed to remain static over time, these can be used as input to the forward model when evaluating model performance on the test set. Hence, for the forward model, say $\mathcal{F}$, $\mathcal{F}$ : $[x_{test}, \hat{z}_{val}] \rightarrow y_{test}$ mapping can be used to evaluate model performance. This allows us to eliminate overfitting that would have happened had $\hat{z}_{test}$ been used since it would have been computed from $[x_{test}, y_{test}]$.

\section{Evaluation Metric Definitions}

\textbf{NSE}

NSE (Nash-Sutcliff Efficiency) is a measure similar to $R^2$ and is used to measure prediction performance in time-series hydrological models. $Q$ refers to streamflow at time step $i$. In our study, we also evaluate the static variable estimates using the same formula - which is equivalent to $R^2$ score.

\begin{equation*}
    \text{NSE} = 1 - \frac{\sum_i^N (Q_i - \hat{Q}_i)}{Q_i - \bar{Q}_i }
\end{equation*}

We also evaluate the quality of uncertainty estimates. A detailed definition of these metrics can be found in this paper \cite{psaros2023uncertainty}.

\textbf{Calibration Error} 

We estimate the calibration error in the prediction distribution using root mean squared calibration error given as,

\begin{equation*}
    \text{Calibration Error} = \sqrt{  \frac{1}{N_p} \sum_j^{N_p} [p_j - \frac{1}{N}\sum_i^N \mathds{1}\{ u_i \le \hat{u}(x_i)_{p_j}  \}]  }
\end{equation*}

Here, $p_j$ refers to different percentiles for which the $u$ observations are compared against the percentiles from the predicted distribution $\hat{u}$. 

\textbf{Dispersion}

Dispersion has been proposed as a measure for evaluating statistical consistency of uncertainty estimates. A more disperse model is more robust to distributional shifts \cite{psaros2023uncertainty}.

\begin{equation*}
    \text{Dispersion} = \frac{SD_{\sigma}}{\mu_{\sigma}}
\end{equation*}

In the dispersion formula, SD and $\mu$ refer to the standard deviation and mean of the epistemic uncertainty, $\sigma$.

\textbf{Coverage Rate}

Here, the coverage rate evaluates the proportion of times the observed value is being captured by the predicted confidence interval bounds.

\begin{equation*}
     \text{coverage rate} =  \frac{\sum_i^N \mathds{1} (z_i \in [\mu_{z_i} - \sigma_{z_i}, \mu_{z_i} + \sigma_{z_i}])}{N}
\end{equation*}

Here, we evaluate the number of times the observed value $z_i$ lies within the confidence bounds defined as $(\mu_{z_i} - \sigma_{z_i}, \mu_{z_i} + \sigma_{z_i})$.

\end{document}